\documentclass[sigconf,natbib=true]{acmart}

\usepackage{multirow}
\usepackage{float} 
\usepackage{amsmath}
\usepackage[ruled,linesnumbered]{algorithm2e}
\usepackage{graphicx}
\usepackage{subfigure}
\usepackage{tablefootnote}
\usepackage[flushleft]{threeparttable}
\usepackage{diagbox}
\usepackage{booktabs}
\usepackage{siunitx}
\usepackage{footmisc}
\usepackage{footnote}
\usepackage{enumitem}
\usepackage[normalem]{ulem}
\useunder{\uline}{\ul}{}

\begin{document}

\title{BLADE: Enhancing Black-box Large Language Models with Small Domain-Specific Models}

\author{Haitao Li}
\affiliation{DCST, Tsinghua University}
\affiliation{Quan Cheng Laboratory}
\email{liht22@mails.tsinghua.edu.cn}

\author{Qingyao Ai}
\affiliation{DCST, Tsinghua University}
\authornote{Corresponding author}
\affiliation{Quan Cheng Laboratory}
\email{aiqy@tsinghua.edu.cn}

\author{Jia Chen}
\affiliation{DCST, Tsinghua University}
\affiliation{Quan Cheng Laboratory}
\email{chenjia0831@gmail.com}

\author{Qian Dong}
\affiliation{DCST, Tsinghua University}
\affiliation{Quan Cheng Laboratory}
\email{dq22@mails.tsinghua.edu.cn}

\author{Zhijing Wu}
\affiliation{Beijing Institute of Technology}
\email{zhijingwu.bit.edu.cn}

\author{Yiqun Liu}

\affiliation{DCST, Tsinghua University}
\affiliation{Zhongguancun Laboratory}
\email{yiqunliu@tsinghua.edu.cn}

\author{Chong Chen}
\affiliation{Huawei Cloud BU}
\email{chenchong55@huawei.com}

\author{Qi Tian}
\affiliation{Huawei Cloud BU}
\email{tian.qi1@huawei.com}

\begin{abstract}
Large Language Models (LLMs) like ChatGPT and GPT-4 are versatile and capable of addressing a diverse range of tasks. However, 
general LLMs, which are developed on open-domain data, may lack the domain-specific knowledge essential for tasks in vertical domains, such as legal, medical, etc. To address this issue, previous approaches either conduct continuous pre-training with domain-specific data or employ retrieval augmentation to support general LLMs. Unfortunately, these strategies are either cost-intensive or unreliable in practical applications. To this end, we present a novel framework named BLADE, which enhances \textbf{B}lack-box \textbf{LA}rge language models with small \textbf{D}omain-sp\textbf{E}cific models. BLADE consists of a black-box LLM and a small domain-specific LM. The small LM preserves domain-specific knowledge and offers specialized insights, while the general LLM contributes robust language comprehension and reasoning capabilities. Specifically, our method involves three steps: 1) pre-training the small LM with domain-specific data, 2) fine-tuning this model using knowledge instruction data, and  3) joint Bayesian optimization of the general LLM and the small LM. Extensive experiments conducted on public legal and medical benchmarks reveal that BLADE significantly outperforms existing approaches. This shows the potential of BLADE as an effective and cost-efficient solution in adapting general LLMs for vertical domains.

\end{abstract}


\maketitle

\section{Introduction}
Recently, large language models (LLMs) have attracted considerable attention in both academia and industry~\cite{openai2023gpt4,zeng2022glm,dong2023aligning}. These models, driven by expansive neural networks and trained on extensive data sets, exhibit remarkable ability in comprehending and generating natural language.
The wide application of LLMs trained with open-domain data, denoted in this paper as \textit{General LLMs}, has profoundly impacted various aspects of daily life and professional environments.

Despite their superior capabilities, large language models often face challenges in vertical domains (e.g., medicine, legal) where in-depth, domain-specific knowledge is crucial~\cite{chalkidis2023chatgpt,cheng2023adapting}. 
For instance, as shown in this paper, ChatGPT exhibits suboptimal performance in Chinese legal question-answering tasks due to its limited knowledge of the Chinese legal system.
Therefore, how to adapt general LLMs for domain-specific applications has become an important problem for the research community~\cite{cheng2023adapting,arefeen2023leancontext}.

Existing methods for adapting general LLMs to specific domains can be broadly divided into two main categories: domain data continuous pre-training and retrieval augmentation. Continuous pre-training involves infusing domain knowledge into general LLMs by training them on a domain-specific corpus~\cite{aharoni2020unsupervised,sachidananda2021efficient}. 
While straightforward, this paradigm requires direct access to large-scale domain data and LLM parameters, which are not available in many conditions. Also, even with access to general LLM parameters and sufficient domain-specific data, directly tuning a general LLM (such as GPT-4) can be prohibitively expensive and poses a risk of overfitting. Aware of these challenges, researchers propose retrieval augmentation as a new paradigm, aiming to enhance general LLMs by leveraging their in-context learning ability~\cite{shi2023replug}. It involves first using a text retriever to find relevant content from the domain corpus, which is then incorporated into the LLM's input to aid in understanding domain-specific knowledge. 
However, there may exist two problems in this paradigm.
First, retrievers primarily rely on exact matches or semantic similarity, lacking inferential capabilities. This limitation means they may not always retrieve documents that fully address specific queries.
Second, the retrieved documents may include irrelevant or misleading information, which could adversely affect the performance of LLMs.


When humans face questions in new domains, besides taking classes (i.e., continuous pre-training) or conducting online searches via platforms like Google (i.e., retrieval augmentation), a more direct and practical approach is to seek advice from experts possessing domain-specific knowledge. 
With this idea in mind, we present BLADE, a novel paradigm where the general LLM is viewed as a black box and the small domain-specific LM ($\#$parameters < 3B) is added as a tuneable module. 
BLADE leverages the superior language comprehension and logical reasoning capabilities of the general LLM, while also incorporating the domain-specific expertise and precision provided by the smaller, domain-focused LM. This approach includes Domain-specific Pretraining (DP) of the smaller LM and introduces two strategies: Knowledge Instruction Tuning (KIT) and Bayesian Prompted Optimization (BPO). Knowledge Instruction Tuning leverages general LLMs to generate pseudo data, which refines the smaller LM, enabling it to generate answers tailored to specific queries. Then, the Bayesian Prompted Optimization aligns the output of small LMs with general LLMs using derivative-free optimization on soft embeddings.
 
To validate the effectiveness of BLADE, we conduct experiments on question-answering tasks in legal and medical fields, which require in-depth knowledge and strong reasoning abilities. The experiments show that BLADE can improve the performance of diverse general LLMs across legal and medical benchmarks. Compared with existing retrieval augmentation methods, the domain-specific small LM can generate more in-depth, comprehensive, and contextually appropriate external knowledge. This capability significantly improves the application of general LLMs in specialized domains. Overall, the principal contributions of this paper can be summarized as follows:

\begin{enumerate} 
\item We introduce BLADE, a new framework for adapting general LLMs to specific domains. This method involves training a smaller, domain-specific Language Model (LM) that aids the general LLM in excelling at tasks within its respective domain.
\item We propose Knowledge Instruction Tuning (KIT) and Bayesian Prompted Optimization (BPO). These strategies enhance the small LM's adaptation to general LLMs, resulting in improved performance.
\item We conduct extensive experiments on public legal and medical benchmarks. Our method significantly outperforms existing approaches that rely on continuous pre-training or retrieval augmentation in both professional fields.
\end{enumerate}

\begin{figure*}[t]
\centering
\includegraphics[width=0.85\textwidth]{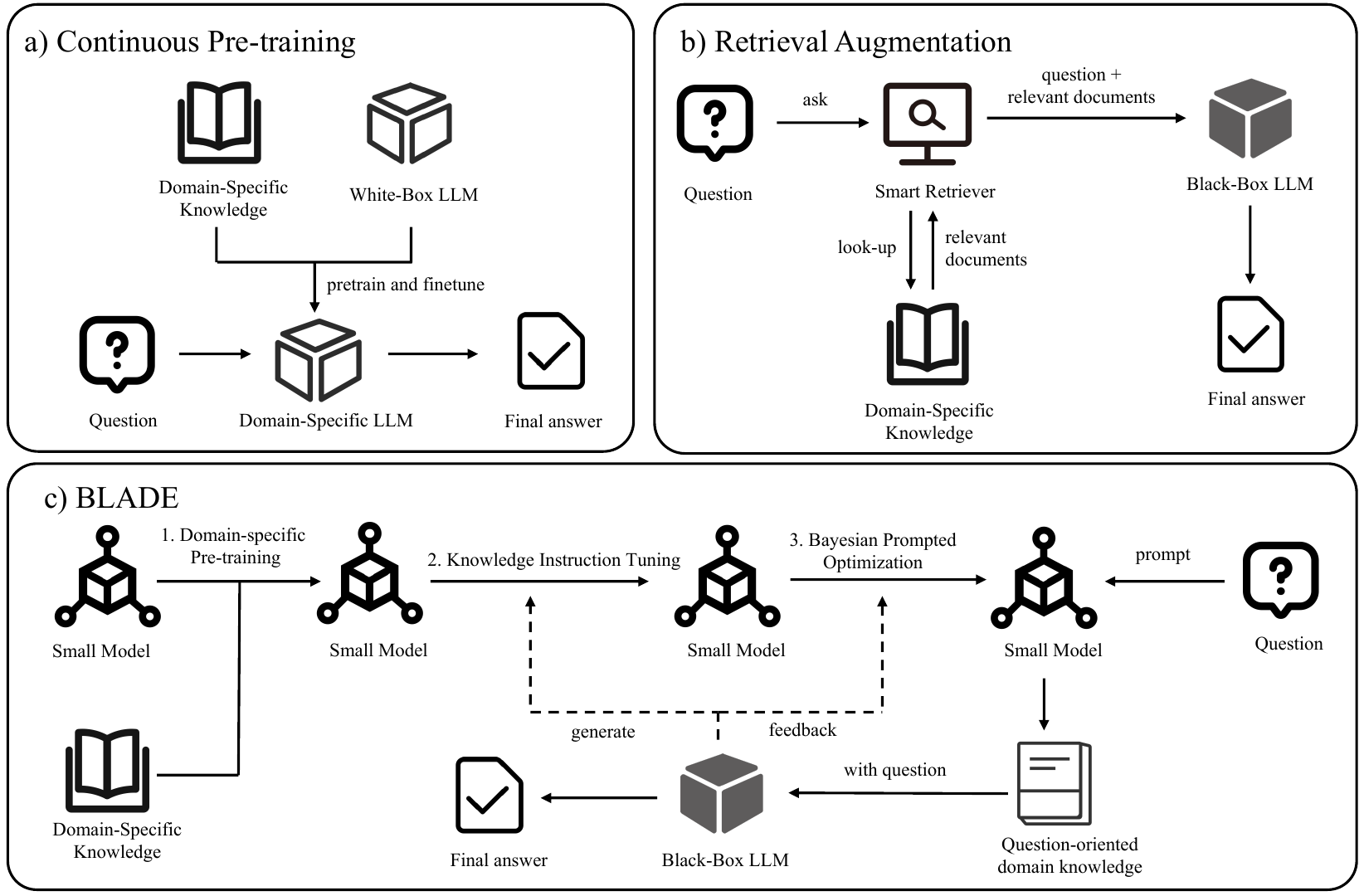}

\caption{Comparison of the workflow of BLADE with existing domain adaptation methods. There are three steps in BLADE: (1) Domain-specific Pre-training imparts domain knowledge to the small LM. (2) Knowledge Instruction Tuning, which enhances the small LM's ability to follow instructions, thereby sharpening its capacity to produce precise, question-specific knowledge. (3) Bayesian Prompted Optimization contributes to aligning the output of small LM with the comprehension of black-box LLM.}
\label{overview}
\vspace{-4mm}
\end{figure*}

\section{Related Work}
\subsection{Large Language Models}
Recently, the emergence of LLMs has attracted substantial attention and research efforts across various fields, primarily owing to their superior linguistic capability.
The foundation for modern LLMs is the Transformer architecture~\cite{vaswani2017attention,chu2024pre}.
which is fundamental in major language models like  BERT~\cite{devlin2018bert} and GPT~\cite{radford2018improving}.
BERT~\cite{devlin2018bert} achieves promising results by pre-training on large corpus and fine-tuning on downstream tasks~\cite{SAILER,li2023constructing,Xieranking}. Similarly, GPT~\cite{radford2018improving} introduces auto-regressive language modeling, generating text sequentially based on previously produced tokens. 
Both BERT and GPT series models require extensive pre-training on a massive corpus to acquire general linguistic knowledge, leveraging tasks such as masked language modeling~\cite{devlin2018bert}, next token prediction~\cite{radford2018improving}, etc.
However, infusing domain-specific knowledge during the pre-training phase of LLMs poses challenges due to the substantial data requirements.
In specific domains, the availability of sufficient data for effective pre-training is often limited, thus constraining LLMs' applicability. 
This limitation is particularly acute in fields with stringent data privacy regulations, such as legal and medical domains.
Another crucial training stage for LLMs is Supervised Fine-Tuning (SFT), which involves training LLMs using task-specific datasets with labeled examples.
This stage adapts the general linguistic knowledge acquired during pre-training to specific tasks, such as sentiment analysis~\cite{dong2021latent}, text classification~\cite{dong2021legal,dong2022incorporating}, and dialogue systems~\cite{ouyang2022training,dong2023i3}.
Despite the effectiveness of SFT, high-quality task-specific annotation is usually costly.
This cost is further amplified by the scale of parameters in LLMs, making the adaptation of LLMs to specific domains during this stage financially prohibitive.
In this paper, we investigate how to leverage existing data to adapt LLMs to specific domains effectively, without the need of additional LLM pre-training or extra annotations.

\subsection{Domain adaptation of LLMs}

The domain adaptation of LLMs is an extensively researched field. Researchers have explored methods such as continuous pre-training, and retrieval augmentation to improve the performance of LLMs in a specific domain~\cite{aharoni2020unsupervised,sachidananda2021efficient,gururangan2020don,shi2023replug,cheng2023adapting}.
The most intuitive approach is continuous pre-training a language model on domain-specific corpora. Previous research has focused on data selection~\cite{aharoni2020unsupervised,gururangan2020don} as well as adjusting or extending tokenizers~\cite{sachidananda2021efficient} to achieve superior performance in the target domain. Despite being effective, fully training these models is often impractical due to extensive computational costs. Additionally, high-quality LLMs can only accessed through the inference API as black boxes.
One possible alternative is to answer domain-specific questions by retrieving relevant information from a specific knowledge base~\cite{borgeaud2022improving,shi2023replug,lewis2020question}. Retrieval augmentation has been shown to be effective in improving performance on various tasks.
For instance, RETRO~\cite{borgeaud2022improving} modifies the model architecture to incorporate retrieved text. Furthermore, REPLUG~\cite{shi2023replug} treats the language model as a black box and enhances it using a retrieval model.
Additionally, recent research has explored substituting traditional document retrievers with large language model generators~\cite{yu2022generate,li2023prompting,sun2022recitation}. In this paper, we focus on employing smaller language models to tackle domain adaptation challenges and further propose methods to adapt the small models to the large ones.

\begin{figure}[h]
\centering
\includegraphics[width=0.8\columnwidth]{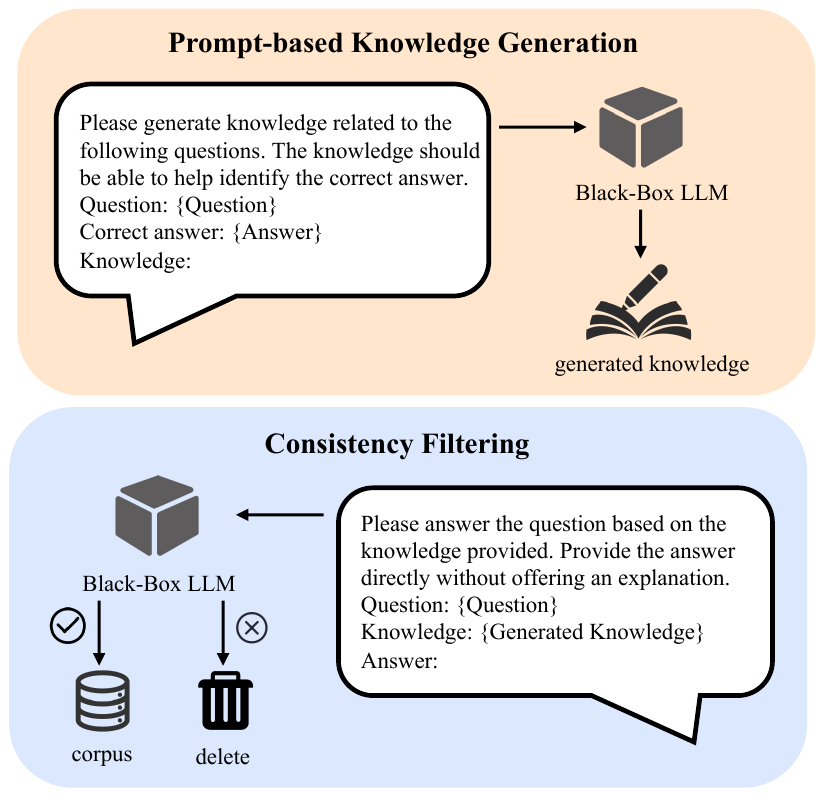}
\vspace{-3mm}
\caption{The process of generating data for Knowledge Instruction Tuning. Only knowledge that can help the black-box LLM correctly answer a question is reserved. }
\label{knowledge}
\vspace{-5mm}
\end{figure}

\section{Method}
In this section, we describe our approach in detail. First, we introduce the overview of BLADE. After that, we elaborate on Domain-specific Pre-training (DP), Knowledge Instruction Tuning (KIT), and Bayesian Prompted Optimization (BPO).

\subsection{Overview}
Figure ~\ref{overview} illustrates the comparison between BLADE and existing domain adaptation methods. Compared to the paradigm of continuous pre-training and retrieval augmentation, BLADE solves domain-specific problems through a collaborative approach between general black-box LLMs and small white-box LMs. General black-box LLMs, such as GPT-4 and GLM-130B, excel in reasoning and inference but often present challenges in terms of cost and feasibility for fine-tuning in downstream applications. Conversely, small white-box LMs may lack sufficient reasoning ability, but can easily update and memorize domain-specific knowledge. 
In the BLADE framework, when faced with a domain-specific query, the small LM initially generates knowledge tailored to the question. Following this, the general LLM synthesizes this information to generate a comprehensive answer.

The training objectives for the small white-box LMs should satisfy two properties. Firstly, these models must effectively memorize domain-specific knowledge. Secondly, they need to effectively communicate with the general black-box LLMs. To achieve this, we introduce two methodologies: Knowledge Instruction Tuning and Bayesian Prompted Optimization. These techniques markedly improve the interaction and collaboration capabilities of the smaller LMs with the general LLMs.

BLADE has several advantages. 
First, small LMs have reduced size and can be easily trained to memorize new knowledge. This separation of knowledge memorization from reasoning capabilities aids in better safeguarding private data. Moreover, unlike the shallow interactions (e.g., inner product) between questions and documents in modern dense retrieval models, the small LM in BLADE generates deeper, question-specific knowledge through intricate token-level cross-attention. We believe that BLADE presents a promising approach for adapting general LLMs to specialized domains, offering a solution that is both effective in performance and cost-efficient.

\begin{figure*}[ht]
\centering
\includegraphics[width=0.8\textwidth]{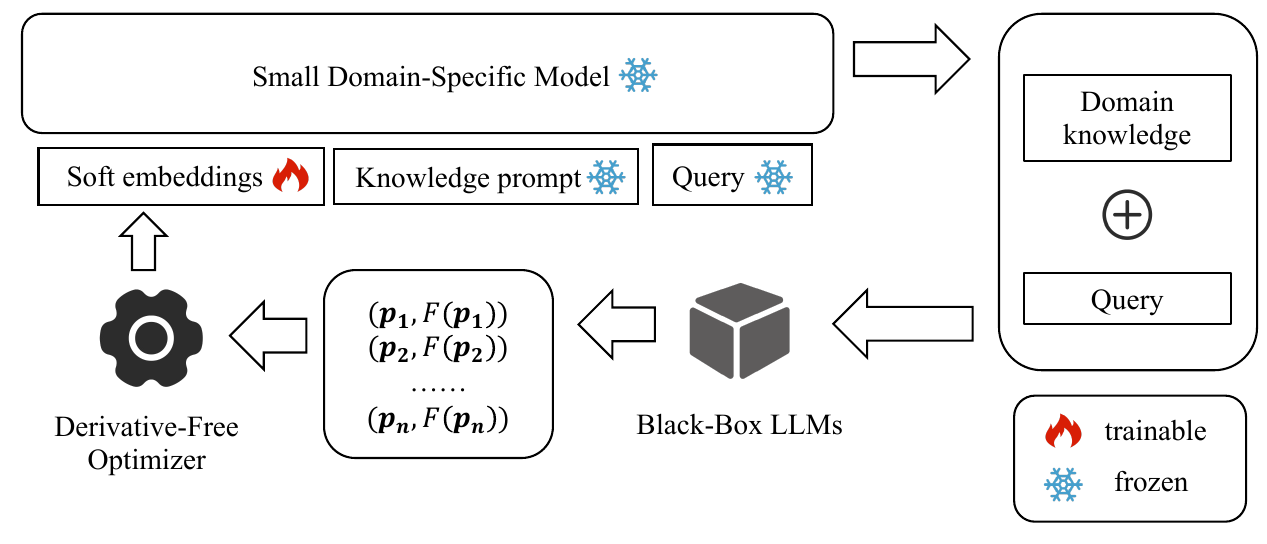}
\caption{Illustration of the Bayesian Prompted Optimization where only soft embeddings are trainable. $F(\boldsymbol{p})$ is the objective score corresponding to soft embedding $\boldsymbol{p}$.
In each iteration, the derivative-free optimizer explores new soft embedding based on previous evaluation scores. The knowledge prompt is consistent with the instruction used in the Prompt-based Knowledge Generation stage.} 
\label{bayesian}
\end{figure*}

\subsection{Domain-specific Pre-training (DP) }

A variety of studies have shown that pre-trained language models implicitly contain substantial knowledge~\cite{wang2022elaboration,liu2021generated,lewis2020retrieval,joshi2020contextualized}. This knowledge can be elicited from the language model through instructional prompts~\cite{liu2021generated}. In light of this, we incorporate domain-specific knowledge via Domain-specific Pre-training (DP).
Specifically, given domain-specific unsupervised text $T = \{t_1,t_2,......,t_n \}$
, we optimize the model by maximizing the following training objective:
$$
G(T)=\sum_{i} \log P\left(t_{i} \mid t_{i-k}, \ldots, t_{i-1} ; \Theta\right),
$$
where $\Theta$ is the parameter of the model. $P$ is the conditional probability of generating the current token based on the previous tokens. After Domain-specific Pre-training, the domain-specific knowledge is effectively encoded into the parameters of the smaller language model.

\subsection{Knowledge Instruction Tuning (KIT) }

While DP enhances small LMs with domain-specific knowledge, its principal purpose is to extend the existing text, rather than actively generating valuable knowledge tailored to the specific queries.
To tackle this limitation, we introduce Knowledge Instruction Tuning (KIT) which focuses on enhancing the instruction-following abilities of the small LM. This process enables the small LM to leverage its inherent knowledge for particular tasks, aligning the model with more practical applications.

To be specific, KIT consists of three components: Prompt-based Knowledge Generation, Consistency Filtering, and Instruction Tuning. Figure ~\ref{knowledge}
show the process of generating data of KIT.
During Prompt-based Knowledge Generation, we combine question-answer pairs from the training dataset with task-specific prompts. Subsequently, a general black-box LLM is employed to generate knowledge that assists in accurately answering the questions. The process entails instructing the model using precise answers and promoting reverse reasoning to deduce the underlying reasons. To enhance the trustworthiness of the generated knowledge, retrieval models can be employed to supply relevant information.
Prompt-based Knowledge Generation is based on the intuition that the general black-box LLM always generates knowledge in a format and style that aligns with its own preferences. Similar to people, we tend to convey our knowledge in a manner that is readily comprehensible to us.

Afterwards, we improve the quality of the generated knowledge by ensuring round-trip consistency. This involves verifying that the general black-box LLM can accurately respond to queries based on the knowledge it produces. Consistency Filtering has been shown to be effective for query generation in QA tasks and various information retrieval tasks~\cite{dai2022promptagator,lewis2021paq}. The instruction prompts are structured as depicted in Figure~\ref{knowledge}.

We filter the generated knowledge and only retain the information that can help the LLM to answer questions correctly. The refined data is then used to fine-tune the small LM, employing the same prompt template as in the Prompt-based Knowledge Generation stage. After KIT, the small LM acquires the ability to generate specific knowledge to the questions.

\subsection{Bayesian Prompted Optimization (BPO)}
In this section, we try to align the output of the small LM with the understanding of the general LLM.
This approach draws inspiration from human society, where pre-collaborative training often leads to better effectiveness. Different from previous work~\cite{chen2023instructzero,sun2022black}, our objective is not to train better black-box general LLMs or to manually craft task-specific prompts. Instead, the primary aim here is to fine-tune the small LM so that it aligns with the general LLM's preferences.

Figure ~\ref{bayesian} illustrates the detailed process of Bayesian Prompted Optimization (BPO) where only soft embeddings are trainable.
To be specific, the primary optimization objective is to enhance the performance of the general LLM$f(\cdot)$ on domain-specific tasks.
Consider an example $(X, Y)$ from the dataset $\mathcal{T}_t$. $k$ represents the domain knowledge that is specific to the query $X$.
In our framework, knowledge $k$ is generated by the small domain-specific LM $g(\cdot)$. $h(\cdot,\cdot)$ is defined as the evaluation metric for output $f(k,X)$ and ground truth $Y$. For example, in multiple-choice tasks, $h(\cdot,\cdot)$ can be accuracy. The optimization objective is to maximize the performance with appropriate knowledge, i.e.,
$$
\max _k \mathbb{E}_{(X, Y) \sim \mathcal{T}_{t}} h(f([k ; X]), Y), \text { s.t. } k=g(X) , 
$$

As discussed in ~\cite{chen2023instructzero}, the above problems are combinatorial optimization with complicated structural constraints. Since $f(\cdot)$ is a black-box model, traditional optimization via backpropagation is not feasible. Therefore, we apply derivative-free optimization to refine the soft prompt $\boldsymbol{p_h}$ on small model $g(\cdot)$. 
Specifically, we concatenate $n$ soft tokens $p_{h_1:h_n} \in  \mathbb{R}^D $  with input queries $X$ as inputs to the small model, facilitating the generation of domain knowledge $k = g(p_{h_1:h_n}, X)$.
Therefore, our objective is to identify the optimal soft prompt:

$$
\boldsymbol{p_h^*} = \text{arg} \max_{\boldsymbol{p_h}\in \mathbb{R}^D}\mathbb{E}_{(X, Y) \sim \mathcal{T}_{t}} h(f([g(\boldsymbol{p_h},X); X]), Y) .
$$

Although the original optimization problem is transformed into a feasible continuous optimization problem, derivative-free black-box optimization remains challenging due to the high dimensionality of the optimized soft prompt. To address this problem, we attempt to optimize a lower dimensional vector $\boldsymbol{p} \in \mathbb{R}^d $ where $d \ll D$ and apply a random projection $A \in \mathbb{R}^{ d \times D}$ to project $\boldsymbol{p}$ into the original space. This is feasible based on (1) 
pre-trained language models possess low intrinsic dimensionality, indicating that the minimal reparameterization required for effective optimization is significantly lower than the full parameter space, as supported by findings in ~\cite{sun2022black,hu2021lora}. (2) According to Johnson-Lindenstrauss Lemma~\cite{kleinberg1997two}, the random projection is distance-preserving. This means that the kernel similarity is consistent before and after the random projection.
Thus, the optimization objective is transformed into the following formula:
$$
\boldsymbol{p^*} = \text{arg} \max_{A\boldsymbol{p}\in \mathbb{R}^D}\mathbb{E}_{(X, Y) \sim \mathcal{T}_{t}} h(f([g(A\boldsymbol{p},X); X]), Y) .
$$

We employ Bayesian optimization (BO) ~\cite{frazier2018tutorial} to handle the above optimization. BO is an effective technique for updating the posterior distribution of the objective function by iteratively incorporating new sample points.
To be specific, we first define the objective function as $F(\boldsymbol{p}) = \mathbb{E}_{(X, Y) \sim \mathcal{T}_{t}} h(f([g(A\boldsymbol{p},X); X]), Y)$. 
Then, we employ the Gaussian Process (GP) as the prior to estimate the distribution of $F(\cdot)$, i.e.,
$$
F(\boldsymbol{p}) \sim GP(\mu, \sigma^2) ,
$$
where $\mu$ is the mean function and $\sigma^2$ is the variance function. This GP can be updated iteratively as the optimization process progresses, incorporating new observations to better approximate the true function and reduce uncertainty w.r.t. its behavior.
For each $\boldsymbol{p_{i}}$, we can obtain a score $F(\boldsymbol{p_{i}})$. Let $D$ denote all collected data in previous BO steps, i.e., $D = \{(\boldsymbol{p_{1}},F(\boldsymbol{p_{1}})),...,(\boldsymbol{p_{n}},F(\boldsymbol{p_{n}}))\}$.
Then the $\mu$ and $\sigma^2$ of the GP can be updated as follows:
$$
\mu(\boldsymbol{p})=c(\boldsymbol{p}, \boldsymbol{P})\left(\boldsymbol{C}+\sigma_{n} ^{2} \boldsymbol{I}\right)^{-1}F(\boldsymbol{P}) ,
$$
$$
\sigma^{2}(\boldsymbol{p})=c(\boldsymbol{p}, \boldsymbol{p})-c(\boldsymbol{p}, \boldsymbol{P})\left(\boldsymbol{C}+\sigma_{n}^{2} \boldsymbol{I}\right)^{-1} c(\boldsymbol{P}, \boldsymbol{p}) ,
$$
where $\boldsymbol{P} = [\boldsymbol{p_1},...,\boldsymbol{p_n}]$, $c(\cdot,\cdot)$ is the covariance function and $\boldsymbol{C}$ is the covariance matrix of $\boldsymbol{P}$. $\sigma_{n}$ is the noise variance. $\boldsymbol{I}$ represents the identity matrix. $\boldsymbol{p_1}$ is randomly initialized. After finishing an iteration, we employ Expected Improvement (EI) to find the next $\boldsymbol{p_{n+1}}$. The Expected Improvement (EI) is a popular acquisition function that balances exploration and exploitation. It quantifies the potential improvement over the current best observed value. Formally, the next soft prompt $\boldsymbol{p_{n+1}}$ is defined as follows:
$$
\boldsymbol{p_{n+1}} \in \text{arg} \max_{\boldsymbol{p} \in \mathbb{R}^d}  \mathbb{E}_{F(\boldsymbol{p}) \sim GP(\mu, \sigma^2)}\left[\max \left\{0, F(\boldsymbol{p})-\max _{i \in[n]} F\left(\boldsymbol{p}_{i}\right)\right\}\right] ,
$$
In practice, we employ an evolutionary search algorithm known as CMA-ES~\cite{hansen2016cma} as the optimization method to identify the most effective soft prompts.
When obtaining $\boldsymbol{p_{n+1}}$, we evaluate performance $F(\boldsymbol{p_{n+1}})$ of BLADE on the training batch. Subsequently, the pair $(\boldsymbol{p_{n+1}},F(\boldsymbol{p_{n+1}})) $is incorporated into the $D$ to update $\mu$ and $\sigma^2$.
This process is performed iteratively until convergence (effectiveness gains less than a threshold for a given number of steps) or reaches the maximum iteration number.

\section{Experiment}
In this section, we first introduce our experimental setup, including
datasets and metrics, baselines, and implementation details. Then,
we report experimental results to demonstrate the effectiveness of
BLADE.

\subsection{Datasets and Metrics}
We conduct extensive experiments with question-answering datasets in the legal and medical domains. Experimental datasets are as follows:

\begin{itemize}[leftmargin=*]
    \item[-] \textbf{JEC-QA}~\cite{zhong2020jec} is the largest Chinese multiple-choice dataset in the legal domain. The legal questions in JEC-QA require high-level reasoning ability and are divided into two types: Knowledge-Driven Questions (KD-questions) and Case-Analysis Questions (CA-questions). There are 26,365 questions in JEC-QA, of which 5,289 of them comprising the test set. It's worth noting that the number of correct options for each question is uncertain.
    
    \item[-] \textbf{CaseHOLD}~\cite{zheng2021does} is an English multiple-choice dataset with the purpose of identifying the relevant holding of a cited case. It contains 53,000+ multiple choice questions with 3,600 questions in the test set.

    \item[-] \textbf{MLEC-QA}~\cite{li2021mlec} is the the largest-scale Chinese multi-choice biomedical QA dataset. This dataset contains five subsets: Clinic(Cli), Stomatology (Sto), Public Health (PH), Traditional Chinese Medicine (TCM), and Traditional Chinese Medicine Combined with Western Medicine (CWM), all of them collected from the National Medical Licensing Examination in China. There are 136,236 questions in MLEC-QA, each presenting five options with one correct answer.
\end{itemize}

We report the zero-shot performance of various LLMs on these datasets. Answers are extracted from model predictions by regular matching. Accuracy serves as the primary evaluation metric.
When the correct answer contains more than one option, the model prediction is considered correct only if it exactly matches the golden answer.

\begin{table*}[ht]
\caption{Zero-shot test accuracy on JEC-QA dataset. BLADE achieves consistent improvements on two types of questions. The gain \% shows the relative improvement of methods compared to the original language model. */** denotes that BLADE performs significantly better than the original language model at $p < 0.05/0.01$ level using the fisher randomization test~\cite{rubin1980randomization}. Best results are marked bold.}
\begin{tabular}{lccccccc}
\hline
\multicolumn{1}{l|}{\multirow{2}{*}{Model}} & \multicolumn{1}{c|}{\multirow{2}{*}{\# Parameters}} & \multicolumn{2}{c|}{KD-questions}                              & \multicolumn{2}{c|}{CA-questions}                             & \multicolumn{2}{c}{All}                   \\ \cline{3-8} 
\multicolumn{1}{l|}{}                       & \multicolumn{1}{c|}{}                               & Original       & \multicolumn{1}{c|}{+BLADE}                   & Original       & \multicolumn{1}{c|}{+BLADE}                  & Original       & +BLADE                   \\ \hline \hline
\multicolumn{8}{l}{\textbf{Legal Specific LLMs}}                                                                                                                                                                                                                                        \\ \hline
\multicolumn{1}{l|}{LaywerLLaMA}            & \multicolumn{1}{c|}{13B}                            & 9.76           & \multicolumn{1}{c|}{12.94**(32.6\%)}                        & 6.05           & \multicolumn{1}{c|}{8.66**(43.1\%)}                       & 7.45           & 10.26**(37.7\%)                        \\
\multicolumn{1}{l|}{LexiLaw}                & \multicolumn{1}{c|}{6B}                             & 15.50          & \multicolumn{1}{c|}{19.63**(26.6\%)}                        & 14.35          & \multicolumn{1}{c|}{18.07**(25.9\%)}                       & 14.78          & 18.66**(26.5\%)                        \\
\multicolumn{1}{l|}{ChatLaw-13B}            & \multicolumn{1}{c|}{13B}                            & 10.32          & \multicolumn{1}{c|}{17.32**(67.8\%)}                        & 5.03           & \multicolumn{1}{c|}{8.08**(60.6\%)}                       & 7.01           & 11.55**(64.8\%)                        \\
\multicolumn{1}{l|}{ChatLaw-33B}            & \multicolumn{1}{c|}{33B}                            & 15.66          & \multicolumn{1}{c|}{21.80**(39.2\%)}                        & 17.01          & \multicolumn{1}{c|}{20.46**(20.3\%)}                       & 16.50          & 20.96**(27.0\%)                        \\ \hline \hline
\multicolumn{8}{l}{\textbf{General LLMs}}                                                                                                                                                                                                                                               \\ \hline
\multicolumn{1}{l|}{ChatGLM-6B}             & \multicolumn{1}{c|}{6B}                             & 17.08          & \multicolumn{1}{c|}{21.19**(24.1\%)}          & 16.64          & \multicolumn{1}{c|}{18.62**(11.9\%)}         & 16.81          & 19.58**(16.5\%)          \\
\multicolumn{1}{l|}{ChatGLM2-6B}            & \multicolumn{1}{c|}{6B}                             & \textbf{27.39} & \multicolumn{1}{c|}{30.81**(12.5\%)}          & 24.09          & \multicolumn{1}{c|}{\textbf{26.34**(9.3\%)}} & \textbf{25.32} & \textbf{28.01**(10.6\%)} \\
\multicolumn{1}{l|}{Qwen-7B-Chat}           & \multicolumn{1}{c|}{7B}                             & 25.78          & \multicolumn{1}{c|}{\textbf{31.26**(21.2\%)}} & \textbf{24.52} & \multicolumn{1}{c|}{25.07*(2.2\%)}           & 24.99          & 27.39**(9.6\%)           \\
\multicolumn{1}{l|}{Baichuan-7B}            & \multicolumn{1}{c|}{7B}                             & 15.31          & \multicolumn{1}{c|}{21.80**(41.4\%)}          & 17.80          & \multicolumn{1}{c|}{21.58**(21.2\%)}         & 16.86          & 21.66**(28.4\%)          \\
\multicolumn{1}{l|}{Baichuan-13B-Chat}      & \multicolumn{1}{c|}{13B}                            & 17.87          & \multicolumn{1}{c|}{23.06**(14.1\%)}          & 19.19          & \multicolumn{1}{c|}{21.71**(13.1\%)}         & 18.69          & 21.21**(13.4\%)          \\
\multicolumn{1}{l|}{Baichuan2-7B-Chat}      & \multicolumn{1}{c|}{7B}                             & 19.23          & \multicolumn{1}{c|}{24.27**(26.2\%)}          & 19.53          & \multicolumn{1}{c|}{21.73**(11.3\%)}         & 19.41          & 22.68**(16.8\%)          \\
\multicolumn{1}{l|}{Baichuan2-13B-Chat}     & \multicolumn{1}{c|}{13B}                            & 25.78          & \multicolumn{1}{c|}{28.29**(9.73\%)}          & 21.80          & \multicolumn{1}{c|}{24.22**(11.1\%)}         & 23.29          & 25.75**(10.5\%)          \\
\multicolumn{1}{l|}{ChatGPT}                & \multicolumn{1}{c|}{-}                              & 20.53          & \multicolumn{1}{c|}{28.45**(38.6\%)}          & 18.70          & \multicolumn{1}{c|}{23.67**(26.6\%)}         & 19.38          & 25.46**(31.3\%)          \\ \hline
\end{tabular}
\label{jec}
\end{table*}

\subsection{Baselines}

We adopt four groups of baselines for comparison: General LLMs, Legal-specific LLMs, Medical-specific LLMs, and Retrieval-augmented LLMs.

\subsubsection{General LLMs} 
We consider a range of multilingual general LLMs: ChatGLM-6B~\cite{du2022glm}, ChatGLM2-6B~\cite{du2022glm}, Baichuan-7B/13B-Chat~\cite{baichuan2023baichuan2}, Baichuan2-7B-Chat/13B-Chat~\cite{baichuan2023baichuan2}, Qwen-7B-Chat~\cite{bai2023qwen}, ChatGPT~\cite{floridi2020gpt}.

\subsubsection{Legal-specific LLMs} 
Legal-specific LLMs are further fine-tuned in the legal corpus to improve the understanding of the law. LaywerLLaMA~\cite{lawyer-llama-report}, LexiLaw~\footnote{\url{https://github.com/CSHaitao/LexiLaw}}, ChatLaw-13B/33B~\cite{cui2023chatlaw}, are considered for the evaluation. The LawyerLLaMA model is developed based on the Chinese-LLaMA-13B~\footnote{\url{https://github.com/ymcui/Chinese-LLaMA-Alpaca}} with a combination of general and legal instructions. LexiLaw is fine-tuned based on ChatGLM-6B with legal datasets.
Additionally, ChatLaw-13B is refined based on Ziya-LLaMA-13B-v1~\cite{fengshenbang}, and ChatLaw-33B is fine-tuned with Anima-33B~\footnote{\url{https://github.com/lyogavin/Anima}}.

\subsubsection{Medical-specific LLMs} 
For Medical-specific LLMs, Taiyi~\cite{Taiyi} and Zhongjing~\cite{yang2023zhongjing} are selected as baseline models.
Zhongjing is developed through a series of pre-training and fine-tuning processes based on the Ziya-LLaMA-13B-v1~\cite{fengshenbang}.
The model's initial stage, referred to as Zhongjing\_base, involves pre-training on an extensive medical corpus. This is followed by Zhongjing\_sft, a version that undergoes multiple rounds of supervised fine-tuning based on Zhongjing\_base. 
Taiyi is a bilingual fine-tuned large language model, specifically designed for diverse biomedical tasks, and is continuously trained on the Qwen-7B.

\subsubsection{Retrieval-augmented LLMs} 
To fully evaluate the effectiveness of our proposed method, we also compare it with Retrieval-augmented LLMs. We utilize the following retrieval models as baseline: BGE-base~\cite{xiao2023c}, M3E-base~\cite{Moka}, GTE-base~\cite{li2023general}, piccolo-base~\footnote{\url{https://huggingface.co/sensenova/piccolo-base-zh}}. These models are advanced text embedding systems capable of converting natural language into dense embeddings suitable for retrieval purposes.
\subsection{Pretraining and Implementation Details}

We implement BLADE with BLOOMZ\_1b7~\cite{workshop2023bloom} as the small LM since BLOOMZ is a multi-language model with diverse sizes.
To construct the Chinese legal pre-training corpus, we collect legal articles, legal books, legal cases, and other resources from official websites~\footnote{\url{https://wenshu.court.gov.cn/}}.  The English legal pre-training corpus is derived from the English division of MultiLegalPile~\cite{niklaus2023multilegalpile}.
In the medical domain, our corpus comprises medical Wikipedia entries and various medical texts. We pre-train the small LMs for 6 epochs using AdamW~\cite{loshchilov2018fixing} optimizer, with a learning rate of 5e-5, batch size of 32, and linear schedule with a warmup ratio of 0.1. 

For KIT, we utilized ChatGPT to generate knowledge data, incorporating three manual demonstrations for in-context learning during the Prompt-based Knowledge Generation stage. We fine-tune the small LM up to 10 epochs using the AdamW~\cite{loshchilov2018fixing} optimizer, with a learning rate of 5e-6, batch size of 32, and linear schedule with warmup ratio 0.1. In the process of BPO, we utilize accuracy as the evaluation metric $h(\cdot,\cdot)$. The number of tokens in soft prompts is set to 5. The entries of the random projection matrix $A$ are drawn from a uniform distribution between [-1, 1]. In the BO process, $A$ is fixed. The dimensionality of $p_l$ is set to 10. The maximum number of iterations is set to 50. All models except ChatGPT use greedy decoding with default settings. For each question, the small LM generates only one piece of relevant knowledge. All the experiments in this work are conducted on 8 NVIDIA Tesla A100 GPUs. To facilitate the reproductivity of our results, we will release the source code for our experiments after the reviewing phase.


\begin{table}[t]
\caption{Overall Zero-shot performance on the English dataset CaseHOLD. */** denotes that BLADE performs significantly better than baselines at $p < 0.05/0.01$ level using the fisher randomization test. Best results are marked bold.}
\begin{tabular}{lcc}
\hline
\multirow{2}{*}{Model} & \multicolumn{2}{c}{CaseHOLD}           \\ \cline{2-3} 
                       & Original       & +BLADE               \\ \hline
ChatGLM2-6B            & 48.03          & 58.19**(21.1\%)         \\
Qwen-7B-Chat           & 54.28          & 57.33**(5.6\%)          \\
Baichuan2-7B-Chat      & 47.53          & 58.69**(23.5\%)         \\
Baichuan2-13B-Chat     & 48.69          & 62.55**(28.5\%)         \\
ChatGPT                & \textbf{62.58} & \textbf{64.78(3.5\%)} \\ \hline
\end{tabular}
\label{english}
\vspace{-3mm}
\end{table}

\begin{table*}[t]
\caption{Overall Zero-shot performance on the medical dataset MLEC-QA. */** denotes that BLADE performs significantly better than baselines at $p < 0.05/0.01$ level using the fisher randomization test. Best results are marked bold.}
\begin{tabular}{lllllllllll}
\hline
\multicolumn{1}{l|}{\multirow{2}{*}{Model}} & \multicolumn{2}{c|}{Cli}                                            & \multicolumn{2}{c|}{CWM}                               & \multicolumn{2}{c|}{PH}                                & \multicolumn{2}{c|}{Sto}                               & \multicolumn{2}{c}{TCM}          \\ \cline{2-11} 
\multicolumn{1}{l|}{}                       & \multicolumn{1}{c}{Original} & \multicolumn{1}{c|}{+BLADE}          & Original       & \multicolumn{1}{l|}{+BLADE}           & Original       & \multicolumn{1}{l|}{+BLADE}          & Original       & \multicolumn{1}{l|}{+BLADE}           & Original       & +BLADE          \\ \hline \hline
\multicolumn{11}{l}{\textbf{Medical Specific LLMs}}                                                                                                                                                                                                                                                                                      \\ \hline
\multicolumn{1}{l|}{Zhongjing\_base}        & 15.58                        & \multicolumn{1}{l|}{35.74**}         & 19.03          & \multicolumn{1}{l|}{37.52**}          & 16.55          & \multicolumn{1}{l|}{36.98**}          & 14.48          & \multicolumn{1}{l|}{34.86**}          & 17.41          & 36.65**         \\
\multicolumn{1}{l|}{Zhongjing\_sft}         & 16.00                        & \multicolumn{1}{l|}{47.92**}         & 18.50          & \multicolumn{1}{l|}{49.64**}          & 15.85          & \multicolumn{1}{l|}{50.24**}          & 15.76          & \multicolumn{1}{l|}{46.12**}          & 18.88          & 47.82**         \\
\multicolumn{1}{l|}{Taiyi}                  & 43.42                        & \multicolumn{1}{l|}{49.72**}         & 32.71          & \multicolumn{1}{l|}{42.99**}          & 35.11          & \multicolumn{1}{l|}{45.63**}          & 31.53          & \multicolumn{1}{l|}{41.77**}          & 32.83          & 43.65**         \\ \hline \hline
\multicolumn{11}{l}{\textbf{General LLMs}}                                                                                                                                                                                                                                                                                               \\ \hline
\multicolumn{1}{l|}{ChatGLM-6B}             & 30.04                        & \multicolumn{1}{l|}{53.42**}         & 30.84          & \multicolumn{1}{l|}{55.06**}          & 30.47          & \multicolumn{1}{l|}{55.66**}          & 27.56          & \multicolumn{1}{l|}{52.24**}          & 32.96          & 53.64**         \\
\multicolumn{1}{l|}{ChatGLM2-6B}            & 48.86                        & \multicolumn{1}{l|}{60.20**}         & 44.82          & \multicolumn{1}{l|}{57.23**}          & 44.39          & \multicolumn{1}{l|}{59.75**}          & 41.77          & \multicolumn{1}{l|}{57.61**}          & 46.12          & 55.72**         \\
\multicolumn{1}{l|}{Qwen-7B-Chat}           & 56.57                        & \multicolumn{1}{l|}{59.78*}          & 52.59          & \multicolumn{1}{l|}{58.20**}          & 52.64          & \multicolumn{1}{l|}{62.26**}          & 49.33          & \multicolumn{1}{l|}{57.39**}          & 51.53          & 56.62**         \\
\multicolumn{1}{l|}{Baichuan-7B}            & 27.80                        & \multicolumn{1}{l|}{54.86**}         & 25.19          & \multicolumn{1}{l|}{56.03**}          & 26.75          & \multicolumn{1}{l|}{58.54**}          & 22.34          & \multicolumn{1}{l|}{50.34**}          & 24.66          & 52.59**         \\
\multicolumn{1}{l|}{Baichuan-13B-Chat}      & 42.17                        & \multicolumn{1}{l|}{58.98**}         & 45.27          & \multicolumn{1}{l|}{56.59**}          & 42.01          & \multicolumn{1}{l|}{61.54**}          & 38.52          & \multicolumn{1}{l|}{56.42**}          & 41.97          & 55.66**         \\
\multicolumn{1}{l|}{Baichuan2-7B-Chat}      & 51.10                        & \multicolumn{1}{l|}{59.99**}         & 51.14          & \multicolumn{1}{l|}{58.69**}          & 50.00          & \multicolumn{1}{l|}{62.45**}          & 45.29          & \multicolumn{1}{l|}{57.61**}          & 51.79          & 56.82**         \\
\multicolumn{1}{l|}{Baichuan2-13B-Chat}      & \textbf{58.98}               & \multicolumn{1}{l|}{\textbf{61.62*}} & \textbf{54.39} & \multicolumn{1}{l|}{\textbf{58.79**}} & \textbf{57.92} & \multicolumn{1}{l|}{\textbf{63.80**}} & \textbf{50.39} & \multicolumn{1}{l|}{\textbf{57.84**}} & \textbf{54.87} & \textbf{57.34*} \\
\multicolumn{1}{l|}{ChatGPT}                & 47.56                        & \multicolumn{1}{l|}{58.92**}         & 38.69          & \multicolumn{1}{l|}{57.91**}          & 47.73          & \multicolumn{1}{l|}{63.37**}          & 43.32          & \multicolumn{1}{l|}{57.58**}          & 36.49          & 56.40**         \\ \hline
\end{tabular}
\label{medical}
\vspace{-3mm}
\end{table*}




\subsection{Experiment Result}
\subsubsection{Main result in legal domain}

Table ~\ref{jec} presents the results from the baselines and BLADE on the JEC-QA dataset. We derive the following observations from the experiment results.
\begin{itemize}[leftmargin=*]
    \item Legal-specific LLMs show relatively poor results. 
    There is even some performance degradation after domain-specific fine-tuning. For example, LexiLaw underperforms compared to ChatGLM-6B. We hypothesize that although continuous tuning can enhance domain knowledge, it also significantly impacts the model's ability in prompt processing. This observation is also in line with Cheng et al~\cite{cheng2023adapting}. Furthermore, it's worth noting that the legal-specific LLMs demonstrate substantial improvement when integrated with BLADE, implying that these models may not be fully leveraging the domain knowledge encoded in their parameters.
    \item BLADE consistently enhances performance across various models. For example, Baichuan-7B achieves 28.4\% performance improvement, while ChatGPT achieves 31.3\% performance improvement. This indicates that BLADE is applicable to diverse language models with different sizes.
    \item Overall, BLADE effectively utilizes domain knowledge without affecting the reasoning ability of the original model.  It has achieved state-of-the-art results on the Chinese legal question-answering dataset, demonstrating its effective use of domain-specific information.
\end{itemize}

We also evaluate the performance of the several best general LLMs on the English dataset CaseHOLD. The performance comparisons are presented in Table~\ref{english}. From the experimental results, we have the following findings:

\begin{itemize}[leftmargin=*]
    \item BLADE consistently shows improvements on the English dataset. Notably, Baichuan2-13B-Chat exhibits a more significant enhancement compared to ChatGLM2-6B and Baichuan2-7B-Chat. This suggests that larger models might derive greater benefits from the knowledge generated by the smaller LM. 
    However, in the context of Chinese datasets, an inverse trend is noticed, implying that performance improvements may also depend on the intrinsic knowledge and in-context learning abilities of the models.
    \item Another interesting observation is that the performance enhancement of ChatGPT on this dataset is not substantially high. This may be because the original training data of ChatGPT has already incorporated a portion of legal pre-training data, thus reducing the additional advantage gained from external knowledge sources.
\end{itemize}

\begin{table*}[t]
\small
\caption{The performance comparison of BLADE and Retrieval-augmented LLMs on JEC-QA. The gain \% shows the relative improvement of methods compared to the original language model.  */** denotes that BLADE performs significantly better than the original language model at $p < 0.05/0.01$ level using the fisher randomization test. The best method in each column is marked in bold. The legal\_pretrain corpus contains the entire contents of legal\_article and legal\_book.}
\begin{tabular}{cc|ccc|ccc}
\hline
\multirow{2}{*}{Retrieval\_model} & \multirow{2}{*}{Corpus} & \multicolumn{3}{c|}{ChatGLM2-6B}                                              & \multicolumn{3}{c}{ChatGPT}                                                    \\ \cline{3-8} 
                                  &                         & KD-questions(\%)         & CA-questions(\%)        & All(\%)                  & KD-questions(\%)         & CA-questions(\%)         & All(\%)                  \\ \hline
-                                 & -                       & 27.39                    & 24.09                   & 25.33                    & 20.53                    & 18.70                    & 19.38                    \\
BGE-base                          & legal\_article          & 28.51*(5.3\%)            & 24.95(3.6\%)            & 26.41*(4.3\%)            & 26.73**(30.2\%)          & 19.73*(5.5\%)            & 22.36(15.4\%)            \\
BGE-base                          & legal\_book             & 27.54(0.5\%)             & 23.49(-2.4\%)           & 25.01(-1.2\%)            & 27.19**(32.4\%)          & 19.55(4.5\%)             & 22.42**(15.7\%)          \\
BGE-base                          & legal\_all              & 30.11**(9.9\%)           & 24.13(0.2\%)            & 26.38*(4.1\%)            & 27.75**(35.2\%)          & 20.54**(9.8\%)           & 23.25**(19.9\%)          \\
GTE-base                          & legal\_article          & 27.09(-1.1\%)            & 23.55(-2.2\%)           & 24.88(-1.8\%)            & 22.15**(7.8\%)           & 19.04(1.8\%)             & 20.21(4.3\%)             \\
GTE-base                          & legal\_book             & 25.58(-6.6\%)            & 22.84(-5.2\%)           & 23.86(-5.7\%)            & 21.90**(6.6\%)           & 19.55(4.5\%)             & 20.43(5.4\%)             \\
GTE-base                          & legal\_all              & 25.43(-7.1\%)            & 23.28(-3.4\%)           & 24.09(-4.9\%)            & 22.25**(8.3\%)           & 19.10(2.1\%)             & 20.28(4.6\%)             \\
M3E-base                          & legal\_article          & 28.55*(4.2\%)            & 24.77(2.8\%)            & 26.19(3.4\%)             & 26.03**(26.7\%)          & 20.58**(10.1\%)          & 22.63**(16.7\%)          \\
M3E-base                          & legal\_book             & 27.74(1.3\%))            & 24.77(2.8\%)            & 25.88(2.2\%)             & 26.28**(28.0\%)          & 20.98**(12.2\%)          & 22.97**(18.5\%)          \\
M3E-base                          & legal\_all              & 30.56**(11.6\%)          & 24.88(3.3\%)            & 27.02*(6.6\%)            & 28.20**(37.3\%)          & 21.19**(13.3\%)          & 23.82**(22.9\%)          \\
piccolo-base                      & legal\_article          & 28.85*(5.3\%)            & 23.46(-2.6\%)           & 25.49(0.6\%)             & 26.53**(29.2\%)          & 20.46**(9.4\%)           & 22.74**(17.3\%)          \\
piccolo-base                      & legal\_book             & 29.20**(6.6\%)           & 23.46(-2.6\%)           & 25.61(1.1\%)             & 26.18**(27.5\%)          & 19.07(1.9\%)             & 21.74**(12.17\%)         \\
piccolo-base                      & legal\_all              & 30.72**(12.1\%)          & 24.37(1.2\%)            & 26.75*(5.6\%)            & 28.29**(37.7\%)          & 20.61**(10.2\%)          & 23.50**(21.3\%)          \\ \hline
\multicolumn{2}{c|}{BLADE}                                  & \textbf{30.81**(12.5\%)} & \textbf{26.34**(9.3\%)} & \textbf{28.01**(10.6\%)} & \textbf{28.45**(38.6\%)} & \textbf{23.67**(26.6\%)} & \textbf{25.46**(31.3\%)} \\ \hline
\end{tabular}
\label{retrieval}
\end{table*}

\begin{table}[]
\caption{Impact of the number of retrieved documents on JEC-QA. The retrieved corpus is legal\_all. Best results are marked bold.}
\begin{tabular}{lcccc}
\hline
Model     & doc\_num & KD-questions & CA-questions & All   \\ \hline
-         & 0        & 27.39        & 24.09        & 25.33 \\
M3E\_base & 1        & 30.56        & 24.88        & 27.02 \\
M3E\_base & 3        & 30.71        & 24.67        & 26.93 \\
M3E\_base & 5        & 30.36        & 25.29        & 27.19 \\
M3E\_base & 7        & 29.75        & 24.28        & 26.33 \\
M3E\_base & 9        & 29.63        & 24.40        & 26.36 \\ \hline
BLADE     & -        & \textbf{30.81}        & \textbf{26.34}        & \textbf{28.01} \\ \hline
\end{tabular}
\label{number}
\end{table}

\subsubsection{Main result in medical domain}
Table \ref{medical} shows the performance of BLADE on the medical domain dataset MLEC-QA. From the experimental results, we have the following findings:

\begin{itemize}[leftmargin=*]
    \item Similar to the legal domain, the Medical-specific LLMs exhibit unsatisfactory performance. The challenge of integrating domain knowledge through continuous training without compromising the original capabilities of LLMs deserves further investigation.
    \item  Both Zhongjing\_base and Zhongjing\_sft originally performed suboptimal in MLEC-QA. Surprisingly, under the guidance of generated knowledge, Zhongjing\_sft showed superior performance than Zhongjing\_base. This may indicate that supervised fine-tuning can improve the LLMs' ability to comprehend external knowledge.
    \item  In the medical domain, BLADE demonstrates consistent performance improvements across all five subsets, with Bacihuan2-13B-Chat achieving the best performance. Overall, BLADE is proven to be able to maintain excellent performance in different evaluation tasks under multiple domains, which underscores its robust applicability in real-world settings.
\end{itemize}

\begin{figure*}[t]
\centering
\includegraphics[width=0.9\textwidth]{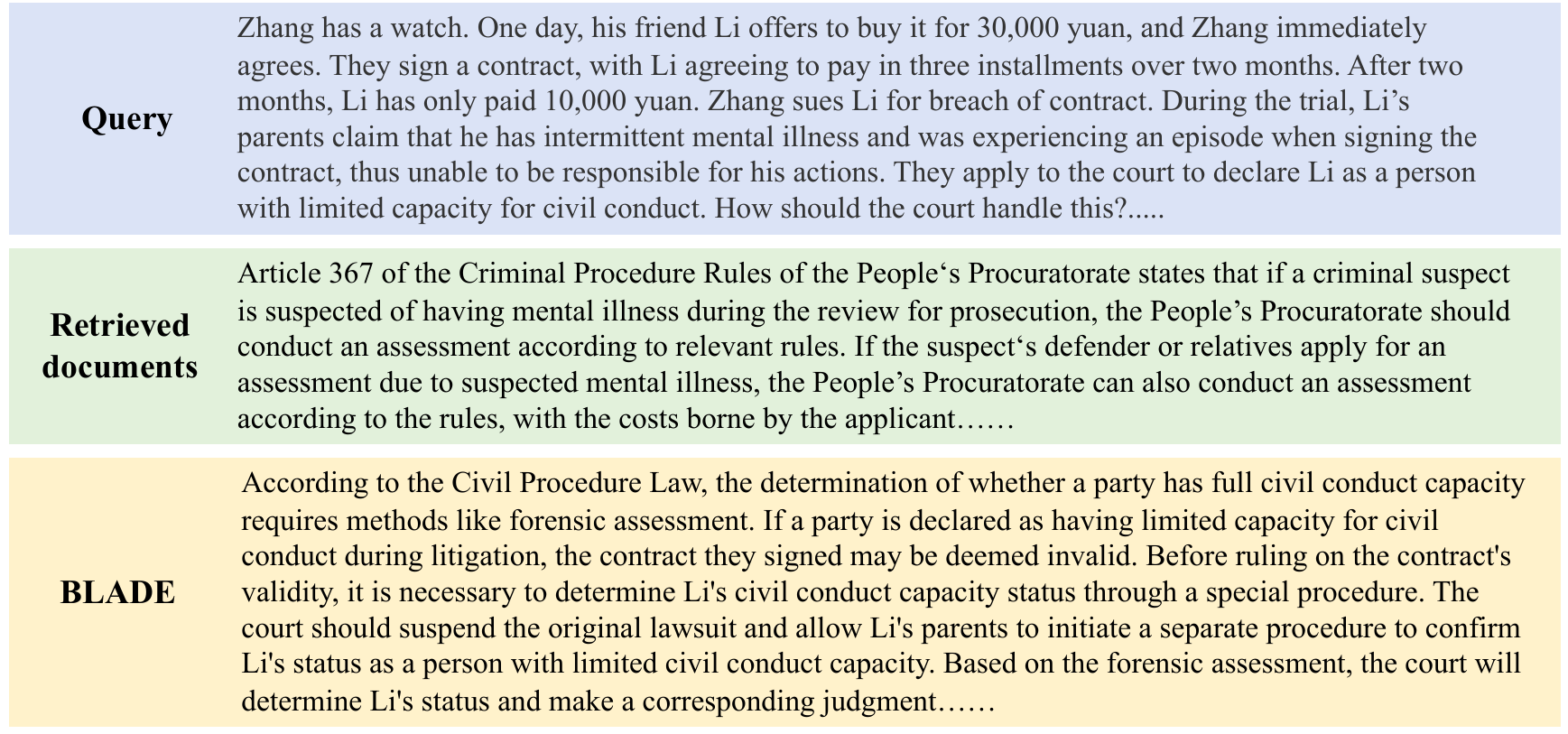}
\vspace{-3mm}
\caption{Comparison of retrieved knowledge with that generated by BLADE.} 
\label{case study}
\end{figure*}

\subsubsection{Comparison with Retrieval Augmented LLMs}
To further explore the effectiveness of BLADE, we compare it to the retrieval augmentation paradigm. More specifically, given a question, the retrieval model first retrieves the most relevant documents from a corpus. 
Then the general LLMs answer the question in the context of the relevant documents. We employ three different legal corpora, including legal\_article, legal\_book, and legal\_all. The legal\_article corpus contains all the Chinese legal provisions.
Legal\_book denotes the National Unified Legal Professional Qualification Examination Counseling Book, which consists of 15 topics and 215 chapters organized in a hierarchical manner. Legal\_all corpus is consistent with the corpus in the DP phase, which contains all documents from legal\_article and legal\_book.

We select ChatGLM2-6B and ChatGPT, which have shown the best results in open-source and closed-source models respectively on the JEC-QA dataset, to conduct the experiment. To ensure a fair comparison, we exclusively use the top-1 document from the retrieval results and employ an identical prompt to BLADE, which also generates only a single document for analysis. The retrieval augmentation methods are implemented using LangChain~\footnote{\url{https://github.com/langchain-ai/langchain}}.
Table ~\ref{retrieval} demonstrates the comparison results. We have the following observations:

\begin{itemize}[leftmargin=*]
    \item Retrieval augmentation is proven to be effective in enhancing the performance of general LLMs in specific domains. However, its effectiveness is significantly influenced by the retrieval model and the corpus. Consequently, not all retrieved knowledge contributes positively to the task at hand.
    \item Knowledge-Driven questions, focusing on the definition and explanation of legal concepts, tend to benefit more from the retrieved knowledge. However, Case-Analysis Questions, involving the analysis of real-life scenarios, may not see significant improvement from retrieved knowledge.
    This reflects the limitations of the retrieval augmentation paradigm, which lacks causal inference ability to identify question-specific knowledge.
    \item Regardless of Knowledge-Driven or Case Analysis questions, BLADE consistently provides stable enhancements and achieves optimal performance. When comparing overall performance with ChatGPT, the most effective retrieval model shows a 22.9\% improvement, while BLADE achieves a notable 31.3\% enhancement
    This success is attributed to our Knowledge Instruction Tuning and Bayesian Prompted Optimization strategy, which effectively supply the crucial knowledge that general LLMs require.
\end{itemize}

We further explore the impact of the number of retrieved documents on the performance of ChagtGLM2-6B. Specifically, we use M3E-base as the retrieval model and legal\_all as the corpus because they achieve the best results in the retrieval augmentation paradigm.
As shown in Table~\ref{number}, when an appropriate number of documents are retrieved, there is a slight performance improvement due to more relevant documents being recalled. However, the performance of ChatGLM2-6B degrades when too many documents are retrieved, probably due to excessive noise introduced by the additional documents. 
In contrast, BLADE achieves the best results by generating only one piece of knowledge, suggesting its proficiency in producing more targeted and refined knowledge.

\begin{table}[t]
\caption{Ablation study on JEC-QA under zero-shot setting. The general LLM is ChatGLM2-6B. Best results are marked bold.}
\begin{tabular}{lccc}
\hline
Small Model   & \multicolumn{1}{l}{KD-questions(\%)} & \multicolumn{1}{l}{CA-questions(\%)} & \multicolumn{1}{l}{All(\%)} \\ \hline
-             & 27.39                                & 24.09                                & 25.33                       \\
BLOOMZ\_1b7   & 26.38                                & 22.40                                & 23.89                       \\
+ DP           & 26.87                                & 23.63                                & 24.85                       \\
+ DP \& KIT      & 28.45                                & 24.89                                & 26.23                       \\
+ DP \& KIT \& BPO & \textbf{30.81}                       & \textbf{26.34}                       & \textbf{28.01}              \\ \hline
\end{tabular}
\label{ablation}
\end{table}

\begin{table}[t]
\caption{Impact of sizes on JEC-QA. The general LLM is ChatGLM2-6B. Best results are marked bold.}
\begin{tabular}{lccc}
\hline
\multicolumn{1}{l}{Small Model} & \multicolumn{1}{l}{KD-questions(\%)} & \multicolumn{1}{l}{CA-questions(\%)} & \multicolumn{1}{l}{All(\%)} \\ \hline
-                               & 27.39                                & 24.09                                & 25.33                       \\
BLOOMZ\_560m                    & 29.05                                & 24.92                                & 26.47                       \\
BLOOMZ\_1b1                     & 29.80                                & 25.52                                & 27.13                       \\
BLOOMZ\_1b7                     & \textbf{30.81}                                & \textbf{26.34}                                & \textbf{28.01}                       \\ \hline
\end{tabular}
\label{Scales}
\end{table}

\subsection{Ablation Studies}

To better illustrate the effectiveness of our approach, we further conduct ablation studies on JEC-QA in zero-shot setting. Table \ref{ablation} shows the impact of different strategies. It's noticeable that while Domain-specific Pretraining successfully imparts domain knowledge to the small LM, it falls short in enabling instruction-following capabilities and in generating suitable knowledge, leading to a decrease in performance. With the integration of Knowledge Instruction Tuning, the small LM begins to offer beneficial knowledge. Bayesian Prompted Optimization further enhances the performance. The above experiments verify the effectiveness of each process within our approach.

\subsection{Impact of Sizes}
In this section, we aim to investigate the impact of the small LM's size. We conducted experiments on the JEC-QA dataset, utilizing ChatGLM2-6B as the general model.
Three versions of the small LM, namely BLOOMZ\_560m, BLOOMZ\_1b1, and BLOOMZ\_1b7, were tested, each trained with the same training parameters and datasets. The results are shown in Table ~\ref{Scales}. We can observe that the small model with 560m parameters can also lead to performance gains. As the parameters of the small LM increase, the performance improvement brought by BLADE also increases. This phenomenon could be attributed to larger models' enhanced capability to generate more accurate and reliable knowledge.

\subsection{Case Study}
In this section, we conduct a case study to facilitate a clear understanding of the effectiveness of BLADE. Figure ~\ref{case study} illustrates the comparison of retrieved knowledge retrieved by M3E-base from the legal\_all corpus with the knowledge generated by BLADE.
This question involves the assessment of civil conduct capacity in the context of a contract dispute. 
The appropriate legal procedure involves suspending the ongoing proceedings and initiating a specialized process by Li's parents to affirm Li's status as a person with limited civil capacity.
The retrieval model returns the article about proceedings for people with mental illnesses, which fails to directly address the civil litigation process and the implications of limited civil capacity in contract disputes. 
BLADE's response is more accurate and directly relevant to the question. It correctly identifies the key issue – the civil litigation process concerning the assessment of civil conduct capacity in the context of a contract dispute. This case shows BLADE's strength in providing domain-specific, contextually appropriate responses. The domain-specific LM, trained on nuanced legal knowledge, is adept at interpreting the underlying legal implications of the described events. Therefore, BLADE can effectively bridge the gap between the specific details of an event and the relevant legal principles or precedents.

\section{Conclusion}
This paper proposes BLADE, a new framework for applying general large language models to new domains. At its core, BLADE employs small language models to assimilate and continually update domain-specific knowledge.
 The framework solves problems by realizing collaboration between general large language models and a small domain-specific model. It comprises three main stages: Domain-specific Pre-training, Knowledge Instruction Tuning, and Bayesian Prompted Optimization. Domain-specific Pre-training injects domain-specific knowledge into the small model. Knowledge Instruction Tuning activates the instruction-following capacity of the small model. Bayesian Prompted Optimization facilitates better alignment of the small model with the large model. Through extensive experiments on legal datasets, we find BLADE consistently demonstrates performance improvement across various language models with different sizes. In the future, we will investigate approaches to minimize hallucinations in small models and explore additional methods for joint optimization. A limitation is that our experiments are conducted only in multiple-choice datasets, the feasibility of our approach in generative tasks still deserves further investigation.

\balance
\bibliographystyle{ACM-Reference-Format}
\bibliography{sample-base.bib}
\clearpage
\appendix

\end{document}